# Answering real-world clinical questions using large language model based systems

Yen Sia Low*[1], Michael L. Jackson*[1], Rebecca J. Hyde[1], Robert E. Brown[1], Neil M. Sanghavi[1], Julian D. Baldwin[1], C. William Pike[1], Jananee Muralidharan[1], Gavin Hui[1,2], Natasha Alexander[3], Hadeel Hassan[3], Rahul V. Nene[4], Morgan Pike[5], Courtney J. Pokrzywa[6], Shivam Vedak[7], Adam Paul Yan[3], Dong-han Yao[7], Amy R. Zipursky[3], Christina Dinh[1], Philip Ballentine[1], Dan C. Derieg[1], Vladimir Polony[1], Rehan N. Chawdry[1], Jordan Davies[1], Brigham B. Hyde[1], Nigam H. Shah[1,7], Saurabh Gombar[1,8]

1. Atropos Health, New York NY, USA
2. Department of Medicine, University of California, Los Angeles CA, USA
3. Department of Pediatrics, The Hospital for Sick Children, Toronto ON, Canada
4. Department of Emergency Medicine, University of California, San Diego CA, USA
5. Department of Emergency Medicine, University of Michigan, Ann Arbor MI, USA
6. Department of Surgery, Columbia University, New York NY, USA
7. Center for Biomedical Informatics Research, Stanford University, Stanford CA, USA
8. Department of Pathology, Stanford University, Stanford CA, USA

*Co-first authors

## Abstract

Evidence to guide healthcare decisions is often limited by a lack of relevant and trustworthy literature as well as difficulty in contextualizing existing research for a specific patient. Large language models (LLMs) could potentially address both challenges by either summarizing published literature or generating new studies based on real-world data (RWD). We evaluated the ability of five LLM-based systems in answering 50 clinical questions and had nine independent physicians review the responses for relevance, reliability, and actionability. As it stands, general-purpose LLMs (ChatGPT-4, Claude 3 Opus, Gemini Pro 1.5) rarely produced answers that were deemed relevant and evidence-based (2% - 10%). In contrast, retrieval augmented generation (RAG)-based and agentic LLM systems produced relevant and evidence-based answers for 24% (OpenEvidence) to 58% (ChatRWD) of questions. Only the agentic ChatRWD was able to answer novel questions compared to other LLMs (65% vs. 0-9%). These results suggest that while general-purpose LLMs should not be used as-is, a purpose-built system for evidence summarization based on RAG and one for generating novel evidence working synergistically would improve availability of pertinent evidence for patient care.

## Introduction

Evidence based medicine, the conscientious, explicit, and judicious use of current best evidence in making decisions about the care of individual patients, has been the standard for the last three decades[1]. However, in some specialties less than 20% of daily medical decisions are supported by quality evidence[2]. The gap between the need for evidence and the availability of evidence in care decisions is driven by two issues. First, the clinical trials often lack generalizability[3], to complex patients who often fail to qualify for trials[4,5]. Such evidence gaps between trials and real-world settings motivate the need for timely and relevant real world evidence (RWE) to guide care and treatment decisions[6,7]. Second, even when studies exist, they frequently have conflicting findings or are so numerous it is difficult to summarize them for a given patient[8,9]. As a result physicians are often in the situation where they require either summarized evidence from reliable sources or custom evidence generated directly taking into account the patient in front of them.

LLMs have increasingly been looked upon as potential sources of such evidence, summarizing from prior literature learned during LLM training[10–12]. While LLMs have displayed impressive performance in responding to natural language queries in various medical domains[13,14], they are prone to hallucinating reference materials or treatment guidelines[15,16] and may produce "recommendations" that are satirical or inappropriate[17].

One approach to adapt LLMs for reliable evidence summarization is the use of a retrieval augmented generation (RAG), where an LLM is used to compile information retrieved from curated knowledge sources.[18] A clinician could submit a clinical question to a LLM and receive summaries of relevant publication (trial reports and observational studies) and practice guidelines retrieved from the knowledge base. This approach is used by OpenEvidence (https://www.openevidence.com) to answer clinical queries via RAG using a LLM. However, in such a RAG system, the answers are limited to pre-existing evidence sources.

Alternatively, an approach that would enable an LLM to generate *on-demand* evidence is to use a LLM with an agent, combining a natural language interface with an evidence generation platform with access to medical record data. In such an agentic system, the LLM would serve as a co-pilot to match clinical intent to an underlying purpose-built agent for generating evidence[7,19]. ChatRWD™ (https://www.atroposhealth.com/chatrwd) uses this approach for answering clinical queries via an LLM chat interface. This LLM-driven user interface translates clinical queries into a structured population-intervention-control-outcome (PICO)[20] study design for processing through an agent that can generate *on-demand* RWE in response to the question.

In this study we assessed the ability of OpenEvidence and ChatRWD to answer clinical questions that might arise in the context of care delivery. As a baseline, we also assessed the ability of three "out-of-the-box" LLMs (ChatGPT-4, Claude 3 Opus, and Gemini Pro version 1.5) to answer the same questions. We hypothesized that OpenEvidence and ChatRWD would outperform the LLMs. We further hypothesized that OpenEvidence would perform well on questions where existing evidence was likely present, while ChatRWD would be the only system capable of providing RWE when relevant published data were lacking.

## Methods

Figure 1 outlines the evaluation process. First, we selected 50 questions as the basis for evaluation (Question Selection) then submitted them to ChatGPT-4, Claude 3 Opus, Gemini 1.5 Pro, OpenEvidence, and ChatRWD. All of the LLM-based systems except ChatRWD provided supporting citations that were then checked for hallucinations (Citation Review). Because ChatRWD performs a new study on demand, the intermediate code generated by ChatRWD for patient cohorting was reviewed by trained clinical informaticians (Study Integrity Review). The output of each system was graded according to a standard medical rubric by a panel of clinicians (Clinical Review).

**Fig. 1: Evaluation process**

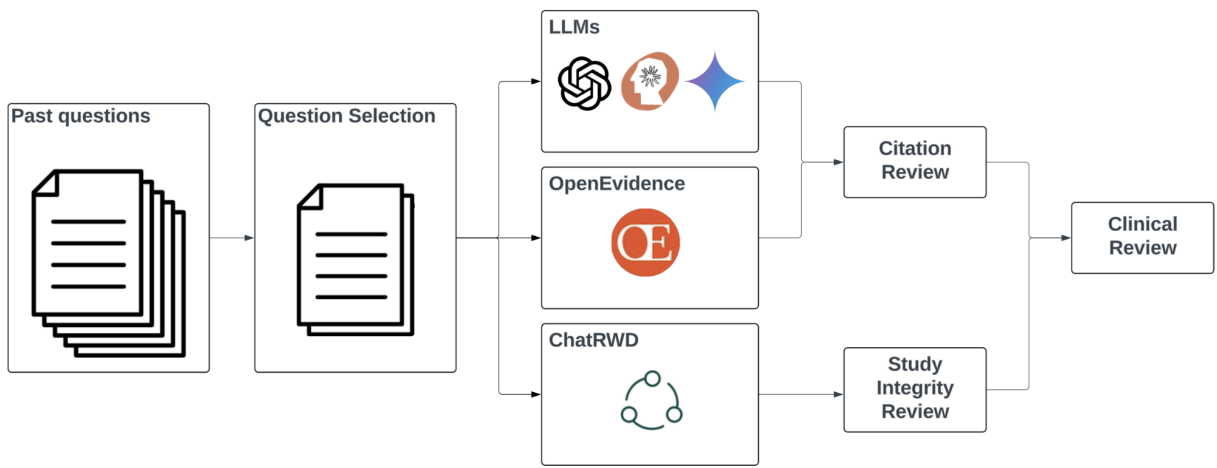

## Question Selection

We selected 50 questions (Table S1) that were either received from physicians requesting additional evidence for clinical decisions or inspired by such questions. All questions met the following following criteria:
- Study is a comparative cohort study;
- All four PICO components can be fully defined;
- Control group is an active comparator rather than absence of treatment;
- Question is not evaluating treatment route, dosage, duration, or line of therapy;
- Intervention does not require washout from prior intervention.

Each question was also assessed for its novelty. This was determined by a consensus vote among three independent reviewers who searched the medical literature for keywords from each question. A question was considered to be novel if the search did not yield any obvious matches for combination of disease indication, treatment, control, and outcome.

## Response Generation

### LLMs

We tested three general-purpose LLMs (ChatGPT-4, Claude 3 Opus, and Gemini Pro version 1.5) to produce answers to the selected clinical questions. OpenAI's ChatGPT[21] is an instruction-tuned pre-trained transformer model designed to produce human-like text in response to natural language instructions. Anthropic's Claude[22] is a suite of AI language and image models trained within a governance framework of pre-specified rules and principles for generative output. Gemini[23] from Google is a suite of natively multimodal AI models designed to interact with and generate multiple data types including text, audio, and images.

We submitted questions to ChatGPT, Claude, and Gemini via their REST APIs. We provided a standardized prompt followed by the clinical question (Fig. S1). The prompt, inspired by previous research[24], directed the model to serve as a helpful assistant with medical expertise. Additionally, in order to facilitate evaluation and understand its reasoning, we specifically asked the models to cite any referenced studies and respond with "I do not know the answer" when that was the case.

### OpenEvidence

OpenEvidence uses an LLM for RAG with different types of medical literature, including PubMed articles and FDA drug labels to answer clinical questions submitted via the web or its API[25]. Using RAG on existing medical literature in this manner reduces the likelihood of hallucinating information, because OpenEvidence can summarize the relevant literature retrieved and present the conclusions to the requester. The final output includes references to the papers identified in the literature search. To evaluate OpenEvidence, we submitted each of the 50 plain English clinical questions through their API and checked the citations provided (Citation Review).

ChatRWD

At its core, ChatRWD includes an agent for cohort selection coupled with statistical analysis to ensure results are not hallucinated. ChatRWD adds four steps to the agent: 1) Chain-of-Thought prompting to convert plain English questions into PICO format, classify the study design, and perform named entity recognition (NER), 2) semantic search of a curated phenotype library, 3) generation of Temporal Query Language (TQL)[26] code to do the cohort-selection and invoke an underlying purpose-built platform for statistical analyses, and 4) summarization of findings from statistical analyses (https://www.atroposhealth.com/chatrwd). Users can confirm and modify the inferred PICO as well as the retrieved phenotypes via a web interface (Fig. S2) before the study is executed. The data source used with ChatRWD (Eversana's Electronic Health Record Integrated Database) consisted of electronic health records of 63 million patients from outpatient and inpatient providers in the United States, including structured medication, laboratory, procedure, and diagnosis data.

## Evaluation of AI-generated responses

### Citation Review

Four of the LLM systems tested (ChatGPT, Claude, Gemini, and OpenEvidence) can cite from the medical literature. We checked the validity of these citations before passing the responses on to our clinical reviewers (Clinical Review). Since OpenEvidence provides links for each citation, we were able to verify the citations by confirming that each link directed to the appropriate study. For the general LLMs, we verified citations by determining if the relevant article could be located on PubMed. We first searched for combinations of author, title, journal, and year of the citation via the PubMed API. For citations without matches, we manually searched PubMed using either the article title alone or the whole reference from the LLM responses. For any citations that were still unmatched, we checked the provided URL from the LLM, if any. Citations still unmatched at that point were considered to be hallucinations.

### Study Integrity Review

ChatRWD does not return citations because the study is run *on-demand* using RWD. Therefore, we conducted a Study Integrity Review instead. ChatRWD involves rule-based generation of TQL code[26] defining a study cohort. For each of the 50 questions, two medical informaticists reviewed the phenotypes and the TQL code for each PICO element to ensure their appropriateness, selecting from one of three grades:

incorrect, not ideal but acceptable, good. The phenotypes were scrutinized to ensure there were no inappropriate inclusions or omissions in order to accurately answer the research question. The study's overall appropriateness and the primary cause of failure to answer the research question if any, was noted.

### Clinical Review

Nine physicians across several specialties graded the responses from all five LLM systems. Before grading, all reviewers underwent training on the use of the standardized rubric (Table S2) using several case studies.

First, they graded along three primary dimensions (response generation, relevance, and evidence quality) to which reviewers had to rate: "yes", "no" or optionally "mixed" for the relevance and evidence quality dimensions (Table S2). We also asked reviewers to evaluate the actionability of each response by selecting "yes" or "no" to indicate whether the response was of sufficient quality to justify or change their practice. Additionally, after reviewing all five responses to each question, reviewers were asked which was the best response.

Because the various LLM systems all have easily identifiable response structures, the physicians were not blinded to the system that generated the answer. The physicians reviewed the content independently and were not able to see the responses of their peers.

### Data analysis

We first aggregated the ratings along the three primary dimensions from the nine clinical reviewers based on a majority vote. From the combination of these aggregated ratings, we then binned each response into one of five exhaustive and mutually exclusive response categories according to the logic shown in Table 1a.

We considered a given response to be actionable if at least five reviewers classified it as high enough quality to justify or change practice. The response receiving the highest number of votes from the nine reviewers, was deemed the 'best', with ties being allocated to both equivalent LLM systems.

We calculated the inter-rater agreement on response category using Krippendorff's Alpha[27] and on the best and actionable metrics using Fleiss' Kappa[28].
For each of the five LLM systems, we assessed an LLM-specific inter-rater reliability of the nine clinical reviewers across 50 questions. We also computed overall inter-rater agreement across all 250 combinations of the 50 clinical questions and five LLM systems.

As a *post hoc* analysis, we examined the concordance between ChatRWD and OpenEvidence as these two LLM systems were most commonly rated as actionable. We hypothesized that OpenEvidence would perform well when relevant published studies exist while ChatRWD would perform best on novel questions without relevant published studies.

## Results

### Relevance, Reliability, and Actionability

The five LLM systems differed widely in their ability to produce relevant, evidence-based results (Table 1a, Fig S3). The LLMs (ChatGPT, Claude, and Gemini) only produced an answer for 58% to 78% of the questions. In comparison, the RAG and agentic systems produced answers for 86% (OpenEvidence) and 94% (ChatRWD) of the questions. When LLMs did produce answers, they rarely gave responses judged to be relevant and evidence-based, meeting this standard for 2% to 10% of the responses. OpenEvidence and ChatRWD, in contrast, respectively produced relevant, evidence-based answers for 24% and 58% of the questions.

Additionally, when considering the stricter criterion for actionability (*i.e.*, of sufficient quality to justify or change clinical practice), the reviewers rarely found the LLM responses (2-4%) to be actionable, while responses from OpenEvidence (30%) and ChatRWD (44%) were more often judged to be actionable (Table 1b). Reviewers most often rated ChatRWD (60%) as providing the best answer, followed by OpenEvidence (46%). None of the LLMs provided the consensus best answer across reviewers.

### Failure Analysis

To understand the causes of poor relevance and poor evidence, we tallied the reasons selected by the clinical reviewers when grading the clinical rubric (Table 2). A major reason for poor relevance was study design mis-specification by ChatRWD (44.7%) and ChatGPT (37.9%, Table 2b). Mis-specification by ChatRWD almost always stemmed from ill-defined phenotypes and sometimes logical errors (*e.g.*, misinterpreting "and" for "or" logic in drug combinations). Some examples of phenotyping errors were when migraine medications included antibiotics (question 9) and when surgery for lower extremities included procedures for upper extremities (question 1).

In contrast, the key reason for the general-purpose LLMs' failure was their frequency of including hallucinated or irrelevant citations, which reviewers identified in 40-80% of relevant responses (Table 2c). Over 40% of all citations from Claude and Gemini could not be located on PubMed, as well as 25.5% of ChatGPT's citations (Table S3). When the answers by LLMs were deemed less relevant, it was often due to minor study mis-specification such as the study population (*e.g.*, ulcerative colitis) not being completely relevant to the question (*e.g.*, Crohn's disease). Of note, OpenEvidence rarely misinterpreted the questions and did not hallucinate citations.

Qualitatively, the reviewers found little value in the responses from the LLMs (Table S4). As one reviewer noted, "Claude, Gemini, and ChatGPT often produce hallucinations and factually incorrect information, necessitating independent verification of all data provided." Two reviewers noted that the advice from the general-purpose LLMs could be potentially fatal. The reviewers were more favorable towards ChatRWD and OpenEvidence: "ChatRWD and OpenEvidence generated the most consistently relevant/robust responses; when both had equal quality responses".

### Inter-rater reliability

Overall, inter-rater reliability across the key metrics was found to be fair (Table S5): Fleiss' Kappa statistics of 0.38 and 0.31 for actionability and best answer respectively, and Krippendorff's Alpha of 0.79 for response categories. Within each LLM system, the inter-rater reliability for response category ranged from poor (ChatRWD: 0.46; OpenEvidence: 0.55) suggesting most subjectivity to good (ChatGPT: 0.85; Gemini: 0.84) with the highest agreement.

### The role of question novelty

To test our hypothesis that ChatRWD would outperform OpenEvidence particularly on novel questions for which ChatRWD can generate new studies, we stratified questions by their novelty and compared the relative performance of ChatRWD and OpenEvidence (Fig. 2, Table 3). Among the novel questions, ChatRWD could produce answers that were actionable (52.2%) as well as answers that were relevant and evidence-based (65.2%). When faced with novel questions, OpenEvidence was rarely able to produce actionable answers (8.7%) or answers that were relevant and evidence based (8.7%).

Conversely, on questions that have existing literature, the comparative gap narrowed (37% relevant and evidence-based, 48.2% actionable by OpenEvidence vs 51.9% relevant and evidence-based, 37.0% actionable by ChatRWD). ChatRWD was more likely to generate answers of varying quality while OpenEvidence would at worst provide partially relevant and partially evidence-based answers (Fig. 2, Table 3, Fig. S4).

This was echoed by clinical reviewers who preferred OpenEvidence when there was existing literature to draw on but noted that ChatRWD was superior when existing literature was absent: "I tended to favor OpenEvidence as its cited studies were both peer-reviewed and generally had more robust study designs (*e.g.,* systematic reviews, RCTs). ChatRWD was most helpful when no studies existed in the published literature for the exact query." For the above reasons, ChatRWD and OpenEvidence were complementary tools. Combined, the actionable rate of answers increased from 30% (OpenEvidence) and 44% (ChatRWD) to 60% for all questions (Table 3).

**Fig. 2: Performance of LLM systems stratified by question novelty**

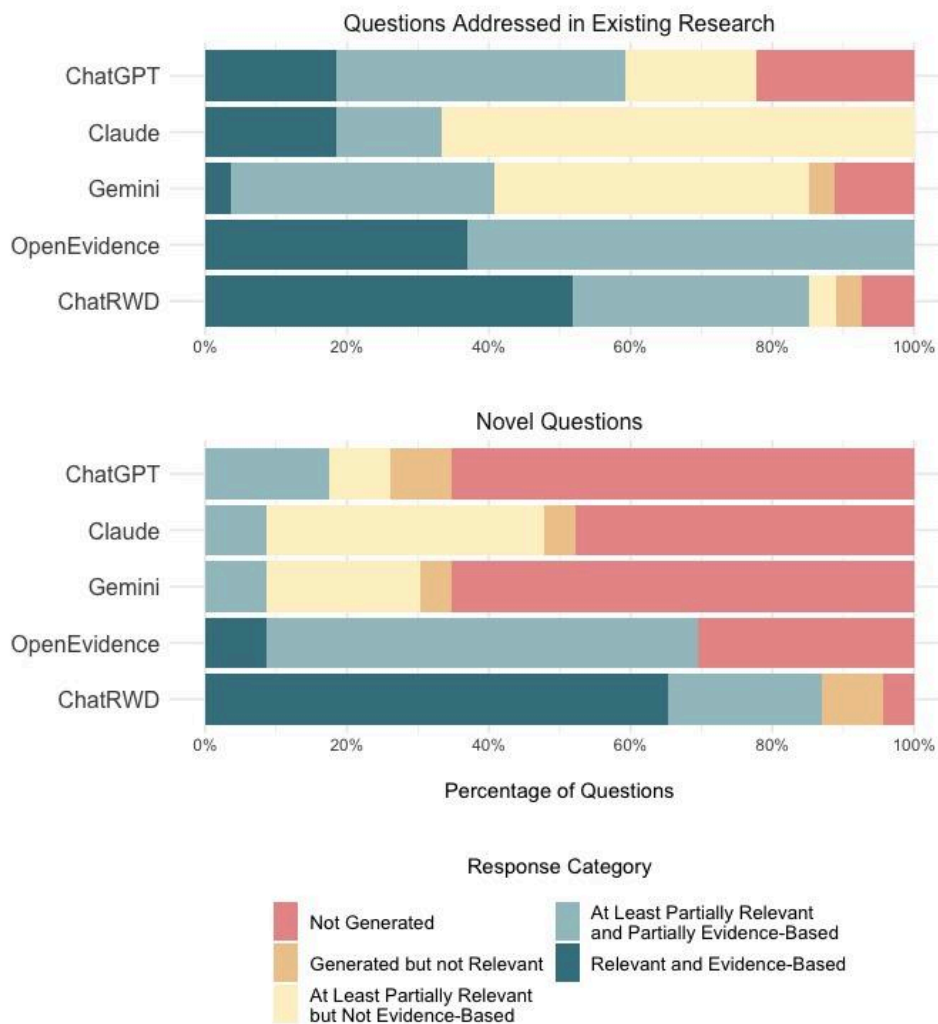

## Discussion

To practice evidence-based medicine it is essential that physicians have rapid access to reliable summarization of trusted published literature as well as have a way to generate

*on-demand* evidence when existing literature does not address the decision at hand. We demonstrated that special-purpose LLM systems when augmented with specialized knowledge (24%, OpenEvidence) or agents (58%, ChatRWD) far outperform off-the-shelf LLMs (2-10%) in producing relevant and evidence-based answers for the clinical questions examined.

The poor performance of general LLMs in this setting of generating RWE is due to several reasons. LLMs hallucinate [29] and use non-credible sources. LLMs cannot appraise sources for relevance, quality and trustworthiness, a critical task for RWE. Particularly harmful is when LLMs are so adept at mimicking the corpora they have been trained on that it becomes difficult to distinguish fact from fiction. Indeed, we observed that probable authors, journals and article titles were often composed together into non-existent citations, making up nearly 40% of citations reported. Further, general-purpose LLMs are not designed for the complex tasks required for RWE generation: study design classification, PICO extraction and clinical NER. Even LLMs that perform well on medical benchmarks such as the USMLE [30] may not generalize across all medical tasks [31]. Finally, LLMs cannot provide responses to novel medical questions whose answer is in content created after the LLMs have been trained. One solution is using RAG to augment LLM with external data sources like the PubMed knowledge base to provide recent evidence sources to draw from, as OpenEvidence has demonstrated.

However, because there is a considerable time lag between asking a clinical question and publishing a comparative study to answer the question, relying only on past studies is inadequate. Given that almost half of our selected questions are novel, there is a huge need to rapidly conduct new studies on demand. These new studies are motivated by patients with specific preexisting conditions and medications who do not qualify for clinical trials. It is precisely for such real-world patients that clinicians struggle to find relevant and high quality evidence that studies have to be performed *on demand* [32].

Thus, it is unsurprising that ChatRWD, specifically designed to conduct comparative studies on demand, outperformed the general LLMs and the RAG-based OpenEvidence particularly for novel questions. This makes OpenEvidence and ChatRWD as complementary tools. OpenEvidence can provide relevant, evidence-based responses to questions that have existing literature, often making use of high-quality sources such as randomized controlled trials and meta-analyses. On the other hand, ChatRWD can generate new evidence for questions that have not previously been studied in the published literature. In combination, these two tools provided relevant, evidence-based answers to 66% of the questions and were deemed actionable 60% of the time.

Several limitations of this study are worth noting. Foremost is that we restricted our clinical questions to those potentially answerable by ChatRWD, meaning treatment comparisons that could be assessed with a cohort study design and used phenotypes already in our phenotype library. Additionally, the novelty rate of the question may not be representative and may affect the relative performance of the various models tested. However, having roughly half of the questions as novel is perhaps a lower bound, given results such as by Darst et al[2]. Third, because each model has a distinctive output format, it was not possible to blind the reviewers to the models. So reviewer preference for a given model can not be controlled for. Fourth, we observed only fair inter-rater reliability across the core measures; this indicates potential variability in results with different reviewer groups. Finally, ChatRWD was allowed access to a single data source due to data rights considerations, and results may be different if run against other RWD sources.

## Conclusions

Pertinent evidence remains difficult to obtain for many patient care decisions. Challenges in obtaining evidence stem from two sources: 1) nearly 80% of care decisions lack high quality evidence due to no specific study being available[2] and 2) difficulty in contextualizing available studies for the specific intricacies of the patient at hand. While LLMs excel at summarizing and contextualizing existing literature, either internalized during training or retrieved from external RAG sources, they cannot perform *new* real-world studies for the exact question at hand unless integrated with an agent to do so. By evaluating the response of five LLMs and having independent reviewers evaluate them for relevance, reliability, and actionability, we demonstrated general-purpose LLMs are not fit for the task of providing evidence for clinical decisions. However, a combination of purpose-built literature retrieval and agent based system to perform *on-demand* studies can do a fair job of surfacing relevant, reliable, and actionable evidence. As these systems continue to improve, it is likely they can be integrated into the physician workflow to enable true evidence-based practice.

## Author contributions

SG, NHS, BH, YL, NS conceived of the study, defined the main outcomes and measures. YL, MJ, RH, and RB drafted the manuscript. NS, JB, SG selected the questions and ran them through ChatRWD. CD ran the questions through OpenEvidence. RH designed the LLM prompts and ran the questions through the general LLMs. RH, SG, medical reviewers searched the literature to corroborate the responses from the LLM systems. SG, JM and NS reviewed the inferred PICO and TQL code. The PICO and TQL evaluation criteria were designed by SG and NS. The medical review rubric was designed by SG, RH, NS, JB and YL. SG trained the clinical reviewers, NA, HH, RN, MP, CP, SV, AY, DY, and AZ who performed the clinician review. RH analyzed and summarized the results from the PICO, TQL, clinical reviews with guidance from YL, MJ, RB, NS and SG. WP, MJ and YL classified the questions by novelty based on literature searches. The phenotype library was adapted for the study by JB, RB, JM, SG, NS, VP, YL, PB. ChatRWD was adapted for the study by NS, YL, RB, JB, RC and JD. A copy of the data for ChatRWD was prepared for this study by DD. All authors reviewed, edited and approved of the final manuscript.

## Acknowledgments

We thank OpenEvidence for technical support and permission to use their platform.

## Competing Interests

# Tables

**Table 1a.** Clinical Review Results: Response category

| Response Category | Dimensions | | | ChatGPT | Claude | Gemini | Open-Evidence | ChatRWD |
| --- | --- | --- | --- | --- | --- | --- | --- | --- |
| | Answer Generated | Relevant | Evidence-based | | | | | |
| Not Generated | No | | | 42% | 22% | 36% | 14% | 6% |
| Generated but not Relevant | Yes | No | | 4% | 2% | 4% | 0% | 6% |
| At Least Partially Relevant, not Evidence-Based | Yes | Mixed or Yes | No | 14% | 54% | 34% | 0% | 2% |
| At Least Partially Relevant, Partially Evidence-Based | Yes | Mixed or Yes | Mixed or Yes | 30% | 12% | 24% | 62% | 28% |
| Relevant & Evidence-Based | Yes | Yes | Yes | 10% | 10% | 2% | 24% | 58% |
| *Total* | | | | *100%* | *100%* | *100%* | *100%* | *100%* |

| Legend | |
| --- | --- |
| *Yes* | 5+ Reviewers responded with "Yes" |
| *Mixed or Yes* | 5+ Reviewers responded with "Yes" or "Mixed" |
| *No* | < 5 Reviewers responded with "Yes" or "Mixed" |
| | Not Applicable |

**Table 1b.** Clinical Review Results: Actionable and Best Answers

| Qualitative Assessment | ChatGPT | Claude | Gemini | Open-Evidence | ChatRWD |
| --- | --- | --- | --- | --- | --- |
| Actionable | 4% | 4% | 2% | 30% | 44% |
| Best | 0% | 0% | 0% | 46% | 60% |

**Table 2a.** Clinical Review Detail: Response generation

|  | ChatGPT | Claude | Gemini | Open-Evidence | ChatRWD |
|---|---|---|---|---|---|
| **Possible Questions:** | **50** | **50** | **50** | **50** | **50** |
| Response Generated | 58.0% | 78.0% | 64.0% | 86.0% | 94.0% |
| No Response Generated | 42.0% | 22.0% | 36.0% | 14.0% | 6.0% |

**Table 2b.** Clinical Review Detail: Relevance, among generated responses

|  | ChatGPT | Claude | Gemini | Open-Evidence | ChatRWD |
|---|---|---|---|---|---|
| **Generated Responses:** | **29** | **39** | **32** | **43** | **47** |
| Relevant | 27.6% | 76.9% | 46.9% | 27.9% | 74.5% |
| Partially Relevant* | 65.5% | 20.5% | 15.6% | 72.1% | 19.1% |
| Not Relevant | 6.9% | 2.6% | 6.3% | 0.0% | 6.4% |
| **Relevance Issues** |  |  |  |  |  |
| Study mis-specification | 37.9% | 10.3% | 9.4% | 7.0% | 44.7% |
|   Major | 3.4% | 0.0% | 0.0% | 0.0% | 19.1% |
|   Minor | 34.5% | 10.3% | 9.4% | 7.0% | 25.5% |
| Incomplete answer | 3.4% | 2.6% | 6.3% | 4.7% | 4.3% |

* No agreement (5+ responses) on "yes" or "no"

**Table 2c.** Clinical Review Detail: Evidence quality, among relevant or partially relevant responses

|  | ChatGPT | Claude | Gemini | Open-Evidence | ChatRWD |
|---|---|---|---|---|---|
| **Relevant or Partially Relevant Responses:** | **27** | **38** | **30** | **43** | **44** |
| Evidence-Based | 51.9% | 18.4% | 6.7% | 86.0% | 84.1% |
| Partially Evidence-Based* | 22.2% | 10.5% | 36.7% | 14.0% | 13.6% |
| Not Evidence-Based | 25.9% | 71.1% | 56.7% | 0.0% | 2.3% |
| **Evidence Issues** |  |  |  |  |  |
| Insufficient Power | - | - | - | - | 4.5% |
| Citation Hallucinated and/or Irrelevant | 40.7% | 78.9% | 80.0% | 2.3% | - |

* No agreement (5+ responses) on "yes" or "no"

**Table 3.** OpenEvidence and ChatRWD performance, separately and combined

|  | OpenEvidence | ChatRWD | Combined |
|---|---|---|---|
| **All Questions** | | | |
| Relevant & Evidence-Based | 24.0% | 58.0% | 66.0% |
| Actionable | 30.0% | 44.0% | 60.0% |
| **Questions Addressed in Existing Research** | | | |
| Relevant & Evidence-Based | 37.0% | 51.9% | 63.0% |
| Actionable | 48.2% | 37.0% | 66.7% |
| **Novel Questions** | | | |
| Relevant & Evidence-Based | 8.7% | 65.2% | 69.6% |
| Actionable | 8.7% | 52.2% | 52.2% |

## Supplemental Tables

**Table S1.** List of questions evaluated and their characteristics

| ID | Novel | Question Text |
|----|-------|---------------|
| 1 | Yes | What is the effect of Estrogen v Progesterone Oral Contraceptives on Post-Operative DVT Incidence in Patients Undergoing Common Lower Extremity Orthopedic Procedures? |
| 2 | No | For male patients over 50 who have Chron's disease is there a difference in hospitalizations over 1 year in those who have ustekinumab vs infliximab? |
| 3 | Yes | For female patients under 35 who have Ulcerative colitis is there a difference in hospitalizations over 1 year in those who have adalimumab vs infliximab? |
| 4 | Yes | Is there a difference in BMI for patients with migraine who receive eptinezumab vs botox therapy? |
| 5 | Yes | For all patients with IBD how does hospitalization differ between those that receive etanercept vs TNF-monoclonals? |
| 6 | Yes | Is there a difference in BMI for patients with migraine who receive gepant therapy vs eptinezumab? |
| 7 | Yes | For patients with autoimmune conditions does the rate of new tinnitis development differ if they received anti-TNF therapy vs other immunotherapy? |
| 8 | No | From one year of starting antipsychotic medications does the number of ER visits differ in patients receiving first generation vs other antipsychotic medicine |
| 9 | No | Are migraine specific medications more effective at reducing the number of ER visits over a year compared non-specific analgesics? |
| 10 | No | Does BMI change more after Estrogen v Progesterone Contraceptives? |
| 11 | Yes | For patients who undergo valve repair surgeries does the rate of MACE differ in those who have statin use vs those who receive Ezetimibe? |
| 12 | No | Is hemoglobin A1C different in diabetic patients who have prolonged steroid use vs those who have prolonged use of non-steroid analgesics? |
| 13 | No | Do GLP1-RA inhibitors result in lower hospitalization rates for heart failure compared to other oral hypoglemics in diabetes patients? |
| 14 | No | Are anti-TNF monoclonal antibodies more effective in preventing major adverse cardiac events compared to traditional DMARDs in patients with rheumatoid arthritis? |
| 15 | Yes | Do patients with hypothroidism who are on Levothyroxine have a lower risk of MACE compared to those who are not on Liothyronine? |
| 16 | Yes | For patients with hypothyroidism does GLP1-RA or bariatric surgery lead to a difference in emergency visits over a year? |
| 17 | Yes | Does length of stay after laprascopic cholecystectomy change between patients on GLP1-RA and those on SGLT-2 Inhibitors? |
| 18 | No | Is the use of proton pump inhibitors more effective in preventing recurrent hospital admissions for upper gastrointestinal bleeding compared to H2 receptor antagonists in patients with a history of bleeding? |

| ID | Novel | Question Text |
|---|---|---|
| 19 | No | How do post surgical infection rates differ between open cholecystectomy and laproscopic cholecystectomy in diabetic patients? |
| 20 | No | Are patients with depression on SSRI therapy less likely to visit the emergency room for psychiatric episodes compared to those on other antidepressants? |
| 21 | Yes | Does length of stay after laprascopic cholecystectomy change between patients on statins and those on niacin? |
| 22 | No | For patients with lung cancer who receive osimertinib alone vs osimertinib plus a PD-L1 antagonist is there a difference in survival? |
| 23 | No | For patients with lung cancer who recieve osimertinib vs other EGFR targetting drugs is there a difference in survival? |
| 24 | Yes | For patients with lung cancer who receive osimertinib alone vs alectinib plus a PD-1L antagonist is there a difference in survival? |
| 25 | Yes | For patients who have OSA after a sleep study is there a difference in depression development in those who have OSA surgery vs those who receive GLP-1 drugs? |
| 26 | Yes | In patients with OSA is there a difference in MACE outcomes for those who receive GLP1-RA vs those who receive metformin? |
| 27 | Yes | In patients with OSA is there a difference in MACE outcomes in those who have liraglutide vs those who receive phenteramine or topiramate |
| 28 | Yes | In patients with OSA is there a difference in MACE outcomes in those that receive OSA surgery vs those who receive liraglutide and no OSA surgery? |
| 29 | Yes | What is the effect of Estrogen v Progesterone Oral Contraceptives on Incidence in patients with any lower extremity fracture? |
| 30 | Yes | For all cancers is there an improved survival benefit win patients who receive Radiation Therapy alone or Radiation Therapy + PDL1 inhibitors? |
| 31 | No | In patients with metastatic Colorectal cancer is there a survival benefit in those that receive bevacizumab vs those that receive cetuximab? |
| 32 | No | Is survival better for patients who have a CABG that receive GLP1-RA vs those that receive other oral hypoglycemic agents? |
| 33 | No | For patients with with Rheumatoid arthritis is there a difference in opioid use in those that receive non-biologic DMARDs vs those that receive Biologic DMARDS? |
| 34 | No | For patients with Rheumatoid arthritis is there a difference in infections in patients that use non-biologic DMARDs vs those that use rituximab? |
| 35 | No | For patients with Rheumatoid arthritis is there a difference in infections in patients that use non-biologic DMARDs vs those that use rituximab? |
| 36 | Yes | For patients with with Rheumatoid arthritis is there a difference in prednisone use in those that receive non-biologic DMARDs vs those that receive Biologic DMARDS? |
| 37 | No | For diabetic patients who do not yet have CKD is there a difference in the rate to progression of CKD in those that have GLP1-RA vs those who have SGLT-2 inhibitors? |
| 38 | Yes | Is there a difference in emergency room visits for patients with constipation who receive |

| ID | Novel | Question Text |
|---|---|---|
| | | colchcine vs misoprostol? |
| 39 | No | Is there a difference in overall survival post acute MI for patients who receive GLP1-RA vs those receiving other oral hypoglycemics? |
| 40 | Yes | Is there a difference in extra-intestinal joint manifestations of IBD in patients who receive adalimumab vs infliximab? |
| 41 | No | In patients with AF who are on a P2Y receptor blocker is there difference in stroke risk for those that receive rivaroxaban vs those that receive apixaban? |
| 42 | No | In patients with AF who are obese is there difference in stroke risk for those that receive rivaroxaban vs those that receive dabigatran? |
| 43 | No | In obese (BMI over 35) patients who are started on VKA vs a DOAC is there a difference in the overall survival? |
| 44 | Yes | For all patients with ankylosing spondylitis how does hospitalization differ between those that receive etanercept vs TNF-monoclonals? |
| 45 | No | What are the outcomes in patients who receive teprotumumab vs. IV steroids for thyroid eye disease? |
| 46 | No | Does absolute neutrophil count differ in patients who receive filgrastim vs Pegfilgrastim? |
| 47 | Yes | Are there any differences in length of stay for spinal tumor patients having transthoracic corpectomy surgery versus lateral extracavitary corpectomy surgery? |
| 48 | No | Is there a survival difference in patients who receive GLP1-RA following transplant vs those that receive other oral hypoglycemics? |
| 49 | No | In patients with AF who are obese is there difference in stroke risk for those that receive apixaban vs those that receive dabigatran? |
| 50 | No | For patients with CKD receiving Denosumab vs bisphosphonates is there a difference in calcium levels? |

**Table S2.** Clinical rubric

| Rubric Item | ChatRWD Possible Answers | LLM & OpenEvidence Possible Answers |
|---|---|---|
| **1. Response Generation:** Was there a PG generated for review? For LLMs, was there an attempt to answer the question? | **Yes** - PG generated | **Yes** |
| | **No** - no PG generated | **No** - response was "I don't know" or study can't be completed |
| | | **No** - response was nonsense |
| | | **No** - other |
| **2. Relevance:** Did the response explicitly answer the question at hand? | **Yes** - exact match | **Yes** - exact match |
| | **Mixed** - still useful, but not exact | **Mixed** - still useful, but not exact |
| | **No** | **No** |
| **3. Evidence quality:** Was the response backed up with reasonable evidence? | **Yes** | **Yes** - all studies cited exist and are relevant & appropriate |
| | **Mixed** - problem with the study but still useful | **Mixed** - some studies exist and are relevant & appropriate |
| | **No** | **No** - studies do not exist or are not relevant & appropriate |
| | | **No** - no studies cited |
| **4. Actionability** With respect to the question at hand, was this result high quality enough to justify or change your practice? | **Yes** | **Yes** |
| | **No** | **No** |
| **5. Best Response** Best Response? *(select one per clinical question)* | ChatRWD ||
| | ChatGPT ||
| | Claude ||
| | Gemini ||
| | OpenEvidence ||

**Table S3.** Citation review*

| Measurement | ChatGPT | Claude | Gemini | OpenEvidence |
|---|---|---|---|---|
| Average Citations / Response | 1.02 | 1.00 | 1.06 | 3.94 |
| % Citations Authenticated | 74.5% | 54.0% | 52.8% | 100% |
| % Citations Unauthenticated | 25.5% | 46.0% | 47.2% | 0.0% |
| % of Responses with an Authenticated Citation | 54.0% | 48.0% | 34.0% | 100.0% |
| % of Responses with an Unauthenticated Citation | 20.0% | 38.0% | 38.0% | 0.0% |

*ChatRWD did not produce citations and was not included in the citation review

**Table S4.** Clinical reviewers' comments

| | Clinical Reviewers' Comments |
|---|---|
| Reviewer 1 | In general, ChatRWD outperforms other LLMs because its PGs align with the questions asked and it can generate large sample size studies. However, there were a few instances where it fell short in areas such as incorrectly defining interventions like medications and procedures codes.<br><br>Other LLMs often struggle to answer questions directly due to a lack of evidence or trials matching the question. Ranked in order of least hallucinations, open evidence includes excellent studies related to the question and can be helpful when interpreted in the clinical context. There were a few QIDs where Open Evidence was better than ChatRWD because of definition inaccuracies in the PGs. GPT, Claude, and Gemini all hallucinate studies and citations, though Gemini's format and response are particularly difficult to read. One small caveat is that GPT and Claude are great at identifying older landmark trials (like EGFR therapy in lung cancer) and even provided better primary evidence (phase III trial results) than Open Evidence (mostly reviews and meta-analysis). |
| Reviewer 2 | Based on my participation in the study, I found that most models present significant limitations. For example, Claude, Gemini, and ChatGPT often produce hallucinations and factually incorrect information, necessitating independent verification of all data provided, while Open Evidence showed promise for answering direct questions but sometimes used incorrect data and drew inappropriate conclusions. Atropos demonstrated the most potential and provided useful results at times but struggled with complex questions, it could also be improved by incorporating more advanced statistical methods (e.g. adjusting for confounders, multivariate analysis etc) and enhancing transparency in data extraction. |
| Reviewer 3 | Overall I felt that chatGPT, claude, and Gemini were unreliable and cannot be used to assist in clinical decision making. ChatRWD was great at answering specific clinical questions that are oftentimes not present in the medical literature, and I would trust the information provided. Open evidence was useful in providing multiple peer reviewed sources to gather more information about a topic, but often did not have the ability to answer the specific question at hand. |
| Reviewer 4 | 1 - open evidence could potentially displace up to date given its ability to tailor the response to the decision at hand.<br>2- Claude and to a lesser extent Gemini will kill people.<br>3- chatRwd is the hinge to precision medicine and the propensity matching allows for high quality studies to exist where they did not before… but only when the data matches the question at hand. |

| Clinical Reviewers' Comments | |
|---|---|
| Reviewer 5 | In general, I found that when there was existing published literature, OpenEvidence was the most helpful. If there was no existing literature, I found ChatRWD to be the most helpful. ChatGPT was generally not that helpful but there was not a lot of hallucination. The other two LLMs contained too many hallucinations for me to find them reliable. |
| Reviewer 6 | When existing clinical studies exist that explicitly explore the clinical question posed to the LLM, OpenEvidence is best positioned to synthesize the numerous existing studies, and provide a coherent summary of the data including limitations of those studies. When no study explicitly addresses the question posed to the LLM, ChatRWD is best positioned to generate new data to help inform the best clinical action. ChatGPT, Claude and Gemini have no role in the healthcare space as the outputs are not reliably based in real data or evidence. The hallucinated content can be shockingly convincing including the creation of citations with credible authors in the space, or the use of existing studies but with mischaracterization of the results. ChatRWD's strengths are that the queries can be traced, the outputs are highly visual and interpretable, and numerous analyses are provided. Weaknesses of ChatRWD is that often the ICD-10 codes included have incorrect codes, and it typically misses time frames included in prompts such as in the last year. |
| Reviewer 7 | Claude consistently produced confidently hallucinated responses, almost always hallucinating the "perfect" study for the query; this made it the least reliable among the LLMs. ChatGPT was occasionally accurate, but its wide range in consistency and frequent hallucinations made it too unreliable for practical use. Gemini was similar to ChatGPT in reliability, but also utilized variably organized Markdown formatting that made it difficult to parse the actual response. OpenEvidence was very impressive in its ability to complete a comprehensive literature review without hallucinations, as well as overall proper interpretation of the studies in context of the query. ChatRWD was also impressive in its ability to appropriately assign PICO elements based on the query, a usually onerous process given the wide range of diagnosis/medication codes, as well as its reliability in summarizing the PG; however, it occasionally either did not complete its PG at variable stages in the generation process, or incorrectly assigned PICO elements. Overall, ChatRWD and OpenEvidence generated the most consistently relevant/robust responses; when both had equal quality responses, I tended to favor OpenEvidence as its cited studies were both peer-reviewed and generally had more robust study design (e.g., systematic reviews, RCTs). ChatRWD was most helpful when no studies existed in the published literature for the exact query, but its PG notably always had to be interpreted with caution given the possible influence of unseen confounders (inability to truly deduce causal relationship). |
| Reviewer 8 | When reviewing medical literature and addressing niche clinical questions, foundational models such as ChatGPT, Claude, and Gemini suffer from frequent hallucinations and inaccuracies, necessitating extensive verification of every output. This issue significantly reduces the utility of these language models as productivity aids. Although prompt engineering and safeguards may mitigate this limitation, I find these off-the-shelf models currently unreliable for anything beyond basic, general medical information. In contrast, OpenEvidence produces outputs that are essentially free from hallucinations, making it substantially more useful. OpenEvidence was also able to use existing studies to synthesize new (often well-reasoned) answers to novel clinical questions, demonstrating a surprisingly sound "reasoning" and understanding of medical concepts.<br><br>I found ChatRWD to be the best for answering specific clinical questions that would normally not have (and unlikely to ever to have) an existing trial, and was great for "hypothesis generation". When there were no other existing high-quality studies, assuming the LLM-driven PICOT translation of the clinical question worked well, assuming the methodology was sound (inclusion criteria was appropriate, enough overlap with propensity scores of the two groups), and assuming that the statistical magic that happens behind the scenes is correct, then I found that ChatRWD was very impressive and helpful. |

| | Clinical Reviewers' Comments |
|---|---|
| | I think the limitations are that there are a few too many "assumptions" above. Part of the discomfort is in not knowing exactly what is happening with the calculations and logic of combining including and exclusion criteria behind the scenes. I don't know if I'd ever truly actually change my clinical practice (especially for important questions like which biologic medication or chemotherapy to prescribe) based on just a PG. I answered the last question as "would you consider changing your clinical practice" when it came to ChatRWD. There were also some instances where the PG generated seemingly high-quality evidence that contradicted what otherwise looked like high-quality peer reviewed medical literature. |
| Reviewer 9 | The various LLM's used in the study had different approaches to answering the specific clinical questions. In the scenario where it does not find the exact answer the models either will give this answer, add in various related details about the individual medications, make a guess on what the answer would be or in some cases criticize the question e.g. the medication is not used in that setting.<br><br>In cases where the model answers the question it may be a full relevant answer with demonstratable figures and conclusions with references given. In many cases the references are either fictitious (authors, journals, study titles, years) but sound like they would be real. The other more concerning issue is that the models may reference a real study however the outcomes and significance of the results have been misstated or misinterpreted.<br><br>The confidence level of the answers may give a reader the misconception that there is truth there and could potentially lead to change in clinical practice if astute critical appraisal is not implemented.<br><br>Some of the models performed better than others for example CHATGPT was quite unreliable or unhelpful both in terms of not having answers or hallucinations whereas Clinical evidence was most helpful in providing high quality evidence where possible and making a conclusion if available in a trustworthy way. |

**Table S5.** Inter-rater agreement*

| Metric | Statistic | ChatGPT | Claude | Gemini | Open-Evidence | ChatRWD | Overall |
|---|---|---|---|---|---|---|---|
| Response Category | Krippendorff's Alpha | 0.85 (good) | 0.73 (fair) | 0.84 (good) | 0.55 (poor) | 0.46 (poor) | **0.79** (fair) |
| Actionable | Fleiss' Kappa | 0.30 (fair) | 0.31 (fair) | 0.36 (fair) | 0.24 (fair) | 0.24 (fair) | **0.38** (fair) |
| Best | Fleiss' Kappa | | | | | | **0.31** (fair) |

* Good/Fair/Poor thresholds for Alpha and Kappa were respectively guided by https://www.k-alpha.org/methodological-notes and https://pubmed.ncbi.nlm.nih.gov/843571/

# Supplemental Figures

## Fig. S1: Prompt for General LLMs

"You are a helpful assistant with medical expertise. You are assisting doctors with their research questions. When answering questions, if possible, cite any studies referenced and mention the study design (e.g. randomized control trial, case-control study, cohort study). At the end of the answer, please provide an APA style citation including a link to the study. If you do not know the answer, respond with 'I do not know the answer'. Try to answer the following question: {insert question here}"

**Fig. S2**: ChatRWD interactive user interface allows user confirmation and modification

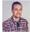

**Fig. S3**: Overall performance of the LLM systems

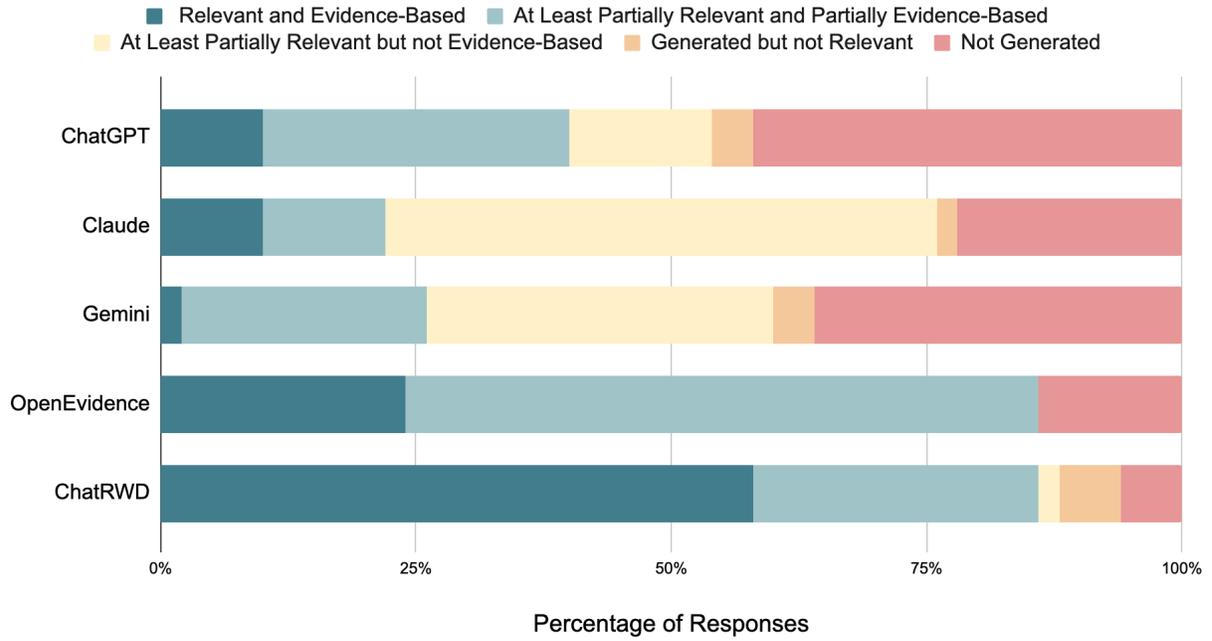

**Fig. S4:** Performance by 5 LLM systems stratified by question and novelty

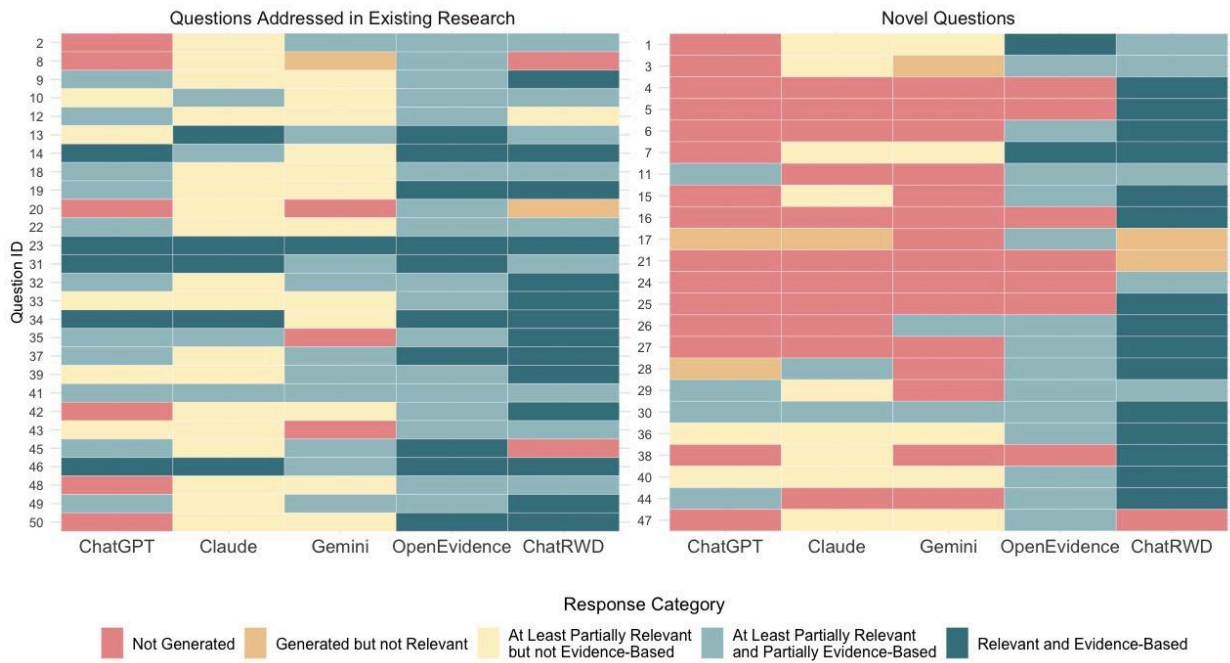